\pdfoutput=1

\documentclass[11pt]{article}

\usepackage{emnlp2021}
\usepackage{times}
\usepackage{latexsym}

\usepackage[T1]{fontenc}

\usepackage[utf8]{inputenc}

\usepackage{microtype}
\usepackage{amsmath}
\usepackage{amsfonts}
\usepackage{amssymb}
\usepackage{mathbbol}
\usepackage[ruled,vlined]{algorithm2e}
\usepackage{newtxmath}
\usepackage{microtype}
\usepackage{booktabs}
\usepackage{stfloats}
\usepackage{graphicx}
\usepackage{xcolor,colortbl}
\usepackage{pifont}
\usepackage{booktabs}
\usepackage{multirow}
\usepackage{soul}
\usepackage{subfigure}

\definecolor{RBRed}{rgb}{0.98,0.88,0.85}
\definecolor{RBBlue}{rgb}{0.81,0.80,0.94}
\definecolor{RBGreen}{rgb}{0.85,0.91,0.84}
\definecolor{DPGreen}{rgb}{0.90,0.45,0.24}
\definecolor{RBBlue2}{rgb}{0.78,0.90,0.90}

\definecolor{L1}{rgb}{1,0.952,0.698}
\definecolor{L2}{rgb}{0.99,0.85,0.46}
\definecolor{L3}{rgb}{0.99,0.698,0.298}
\definecolor{L4}{rgb}{0.992,0.552,0.236}
\definecolor{L5}{rgb}{0.988,0.306,0.1647}

\title{1Cademy at \texttt{}Semeval-2022 Task 1: Investigating the Effectiveness of Multilingual, Multitask, and Language-Agnostic Tricks for the Reverse Dictionary Task}

\author{ 
    Zhiyong Wang \textsuperscript{*} \,  Ge Zhang \textsuperscript{*} \, \textbf{Nineli Lashkarashvili} \\
  University of Colorado Boulder, \,   University of Michigan Ann Arbor, \, San Diego State University \\
  \texttt{zhwa3087@colorado.edu,gezhang@umich.edu,ninelilashkarashvili78@gmail.com}} 

\begin{document}

\maketitle

\begin{abstract}
This paper describes our system for the SemEval2022 task of matching dictionary glosses to word embeddings. We focus on the Reverse Dictionary Track of the competition, which maps multilingual glosses to reconstructed vector representations. More specifically, models convert the input of sentences to three types of embeddings: SGNS, Char, and Electra.
We propose several experiments for applying neural network cells, general multilingual and multitask structures, and language-agnostic tricks to the task. We also provide comparisons over different types of word embeddings and ablation studies to suggest helpful strategies. Our initial transformer-based model achieves relatively low performance. However, trials on different retokenization methodologies indicate improved performance. 
Our proposed Elmo-based monolingual model achieves the highest outcome, and its multitask, and multilingual varieties show competitive results as well.
\end{abstract}

\section{Introduction}
Reverse dictionary Task is defined as word generation based on user descriptions \cite{hill-etal-2016-learning-understand}. 
Following competition rules, pre-trained models and external information should be avoided, and large-scale language models are unsuitable for the task. 
Our paper is devoted to the performance comparison of different neural network structures, multilingual and multitask tricks, and elaborating on language-agnostic or bidirectional structure helpfulness. 
The competition  \cite{mickus-etal-2022-semeval} has significant potential in contributing pretraining process acceleration, low-resource language model development, and commonsense using. 
Furthermore, the task is of high importance for explainable AI and natural language processing since it models direct mapping from human-readable data to machine-readable data.

Known word representation methods using dictionaries, knowledge databases, or glosses have been a common approach for years. 
Related models can be divided into two major groups.
In the former, category methods highly rely on  large-scale model construction.
\citet{levine2019sensebert} develop SenseBert, introducing super-senses from Wordnet \cite{miller1995wordnet} into general Bert model. 
Ernie \cite{sun2019ernie} combines node embeddings from knowledge graph and matched entities to enhance word representations. 
KnowBert \cite{peters-etal-2019-knowledge} subsumes the entity connection and Bert models, which are trained together. 
There are similar research works relevant to the topic \cite{wang2021kepler, wang2020k, yin-etal-2020-sentibert}. 
Still, their models' performances are dependent on the basic large-scale language model trained by sentence samples. 
In the latter group, traditional dependency-based language models learn directly from word dependency and glosses.
They have two major disadvantages: incompatibility with modern language models and relatively low performance  \cite{tissier2017dict2vec, levy2014dependency, wieting2015paraphrase}. 
There is ambiguity about whether recent embeddings and dictionary glosses are mappable from each other.

The paper specifically focuses on progressing utilization of the glosses, different word representations, and languages. 
\textbf{First}, we discuss ablation studies for language-agnostic trick, bidirectional, multilingual, and multitask models and explain the experimental results.      
\textbf{Second}, we apply and analyze different re-tokenization methods. 
\textbf{Finally}, we give instructive conclusions about encoder structures, distinctive word representation relations, and cross-lingual dictionary performance based on our experiment results. We find that \textbf{(1)} transformer-based model performance is inferior to other models for its high complexity, \textbf{(2)} bidirectional models with similar parameter size outperform the unidirectional model because of their better understanding of context-environments even in the low-resource condition, and \textbf{(3)} different word embeddings have a potential relations and can be collaboratively learnt from glosses using  a multitask learning structure.
We make our codes and results publicly available\footnote{\href{https://github.com/ravenouse/Revdict_1Cademy}{https://github.com/ravenouse/Revdict\_1Cademy}}.

\section{Task Description}
\label{task}

The competition, comparing dictionaries and word embeddings, proposes definition modeling \cite{noraset2017definition} and reverse dictionary sub-tracks \cite{hill-etal-2016-learning-understand}. 
These sub-tracks are designed to test the equivalency of dictionary glosses and word embedding representations. 
This paper focuses on the reverse dictionary direction. 
The task refers to word recalling using gloss input and provides word representations that are separately generated by word2vec (SGNS) \cite{mikolov2013efficient}, character \cite{wieting-etal-2016-charagram}, and Electra \cite{clark2020electra} embeddings as training data. 
External data and large-scale language models are strictly restricted from this competition since the models might learn the word embeddings majorly from the sentence samples instead of the dictionary glosses. 
The words matched with the dictionary glosses are hidden in the datasets, implying that dependency-based word representation algorithms cannot be applied directly. 

\section{Methodology}
\label{methodology}
To clarify, we affirm that we only refer to the model structures instead of the trained models when we mention Elmo and MBert in the section and use no external data.
\subsection{Language Model Structure}
\label{language-model}
 
Baseline monolingual models with five distinctive structures were trained: RNN, LSTM, Bi-RNN, Elmo, and Transformer. 

We experiment how bidirectional and different feature generator cell structures help.

\textbf{RNN} is the classical deep learning model dealing with ordinal or sequential data \cite{zaremba2014recurrent}. 
Its major disadvantage is the vanishing and exploding gradient issue.
Nevertheless, the model is fast to converge and works well on smaller sentences. 
Our experiments show that RNN, having similar results to the LSTM-based model, performs slightly better than the transformer-based one.

\textbf{LSTM} is another classical and widely-used feature generator structure in natural language processing. 
The comparison of LSTM-based and RNN-based models can suggest whether vector representation of glosses suffers from the long-term dependencies problem. 
Earlier works \cite{jozefowicz2015empirical} demonstrate that variants of LSTM achieve similar performances in the majority of natural language processing tasks. 
We select the classical LSTM structure for the experiments.

\textbf{Transformer} \cite{vaswani2017attention} is a milestone feature extractor allowing deeper neural network design for natural language processing tasks. 
However, given the much smaller size of the competition data, it performs relatively worse compared to the expectation.

\subsection{Multitask Structure}
\label{multitask}

Although character embedding generation has a similar algorithm to general word embedding methods, it focuses on character representation and is mightier to better tackle the Out Of Vocabulary (\textbf{OOV}) problem. 
We applied Mean Squared Loss (\textbf{MSE Loss}) and Dynamic Weight Averaging (\textbf{DWA}) \cite{liu2019end} as a basic multitask structure for predicting word2vec, Char, and Electra embedding together. 
It achieves competitive performance in both tasks.

\textbf{DWA} \cite{liu2019end} is designed for keeping different tasks converging at the same pace. 
$N$ denotes the number of tasks,  $T$ adjusts the weight-changing sensitivity according to loss difference of the tasks, $L_n(t-1)$ and $r_n(t-1)$ represent the loss and the training speed of task $n$ at $(t-1)$th step. 
$w_i(t)$ is the loss weight of task $i$ at $t$th step. 
The key update equations can be expressed as follows:

\begin{center}
    \begin{equation}
        w_i(t) = \frac{N \, exp(r_i(t-1)/T)}{\sum_n exp(r_i(t-1)/T)}
    \end{equation}
\end{center}

\begin{center}
    \begin{equation}
        r_n(t-1) = \frac{L_n(t-1)}{L_n(t-2)}
    \end{equation}
\end{center}

\subsection{Retokenize Algorithm}
\label{retokenize}

We tried $3$ widely-used retokenization algorithms for vocabulary generation including Byte Pair Encoding (\textbf{BPE}) \cite{sennrich2015neural}, \textbf{WordPiece} Model \cite{schuster2012japanese}, and Unigram Language Model (\textbf{ULM}) \cite{kudo2018subword}. 
BPE is a greedy algorithm that can not model word relation probability successfully. 
WordPiece considers word co-occurrence probability and is influenced by the source data. 
ULM assumes that all subwords are independent and the probability of a subword sequence is the multiple of its element subwords' probability.

\subsection{Multilingual Structure}
\label{multilingual}

We applied two basic multilingual structures for the task: mBert \cite{pires2019multilingual} and adding the language tag. 
\textbf{MBert} has a shared vocabulary for all source languages. 
The results show that mBert can successfully model similar grammar structure, and sentences with similar meanings have akin representations using mBert. 
By applying mBert structure, we can estimate how these important conclusions would work for the reverse vocabularies task. 
We \textbf{add the language tag} as the first token to improve models' ability to separate different languages' representations.

We speculate that language-agnostic representations might aid multilingual models in achieving better performance. 
Residual connection cutting trick proposed by \cite{liu2020improving} was tried, to test how the research findings would work for our specific task.

\begin{figure*}[!tbp]
\begin{minipage}[t]{0.45\linewidth}
\centering
\small
\setlength{\belowcaptionskip}{-0.1cm}  
\includegraphics[height=9.0cm,width=17cm]{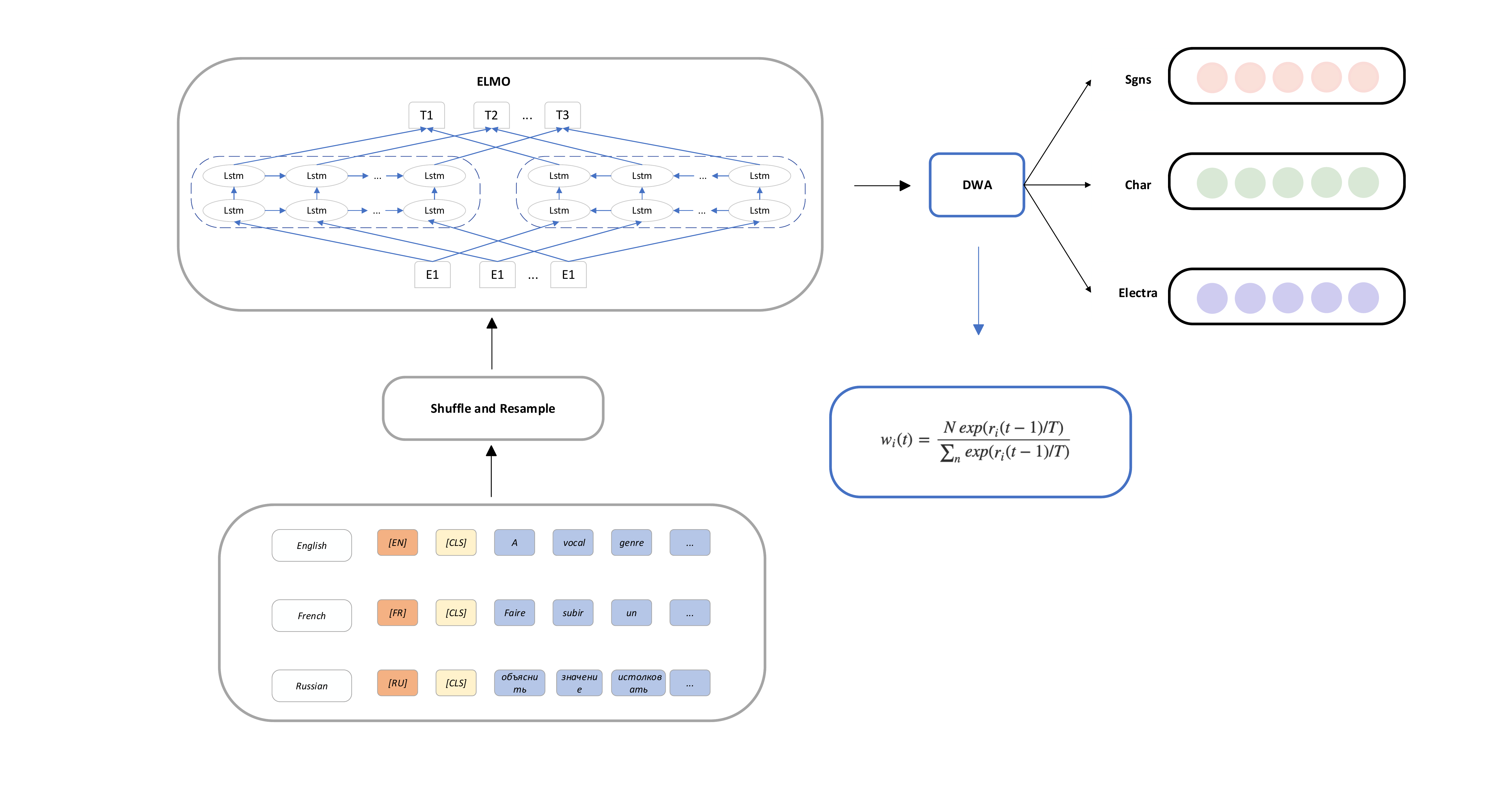}
\end{minipage}
\captionsetup{font={footnotesize}} 
\caption{Sketch Map of the multilingual and multitask Elmo-based model structure.}
\label{fig:layer}
\end{figure*}

\subsection{Selected Model Design}

Following experiment results and ablation studies, our best model is the monolingual Elmo with WordPiece tokenizer. 
The Multitask and multilingual tricks have proved to achieve competitive results with the Elmo language model. 
Adding language tokens achieves a better performance than the plain mBert structure while the Residual Cutting trick does not. 
It implies that the language-specific information is beneficial for the multilingual word representations of the reverse dictionaries task. 
Adding language tokens has demonstrated to help the Elmo-based multilingual model as well.
The most promising multilingual and multitask Elmo-based model structure is shown in \autoref{fig:layer}.

\section{Results and Discussion}
\label{results}
\subsection{Implementation Details}

We apply \textbf{Bidirectional RNN} and \textbf{Elmo} \cite{peters1802deep} models with the same parameter size to find whether bidirectional structure helps.
We selected AdamW \cite{loshchilov2017decoupled} as optimizer. 
All monolingual models share the same hyper-parameters: the number of layers - $4$, the hidden/input size - $256$, and the dropout rate - $0.3$.
WordPiece tokenization was used as the best model design.
We follow \citet{devlin-etal-2019-bert} to set the [CLS] token as the first token for monolingual models. 
We keep the [CLS] token when adding language tokens but set the language token as the first token instead. 

\subsection{Main Results}

Reverse dictionary results are evaluated using three metrics: mean squared error (\textbf{MSE}) between the reconstructed and reference embeddings, cosine similarity (\textbf{COS}) between the reconstructed embedding and the reference embedding, and the cosine-based ranking (\textbf{RANK}) between the reconstructed and reference embeddings, measuring the number of other test items having higher cosine with the reconstructed embedding than with the reference embedding\cite{mickus-etal-2022-semeval}.

\subsubsection{Monolingual Model Performance}

\begin{table*}[ht]
\centering
\resizebox{\textwidth}{!}{%
\begin{tabular}{@{}lccc|ccc|ccc@{}}
\toprule
\textbf{Word Representations}                  & \multicolumn{3}{c}{\cellcolor{RBRed}SGNS} & \multicolumn{3}{c}{\cellcolor{RBGreen}Char} & \multicolumn{3}{c}{\cellcolor{RBBlue}Electra} \\ \midrule Monolingual Models & 
MSE  & COS & RANK    & MSE   & COS    & RANK  & MSE   & COS    & RANK  \\ \midrule
RNN+WordPiece     & 1.000   & 0.249   & 0.310   & 0.158 & 0.778 & 0.442 & \textbf{1.454} & \textbf{0.832} & 0.433    \\
LSTM+WordPiece & 0.990 & 0.228 & 0.375 & 0.148 & 0.791 & 0.458 & 1.491 & 0.831 & 0.449 \\
Transformer+WordPiece & 1.042 & 0.214 & 0.367 & 0.194 & 0.780 & 0.453 & 1.796 & 0.827 & 0.486 \\
BiRNN+WordPiece & 0.989 & 0.221 & 0.395 & 0.150 & 0.791 & 0.454 & 1.483 & 0.832 & 0.449 \\
Elmo+WordPiece & 1.041 & 0.252 & 0.282 & 0.161 & 0.772 & 0.430 & 1.512 & 0.829 & 0.434 \\
Elmo+BPE & 1.037 & 0.250 & \textbf{0.250} & 0.162 & 0.774 & 0.443 & 1.537 & 0.822 & 0.436 \\
Elmo+ULM & 1.022 & \textbf{0.265} & 0.259 & 0.157 & 0.781 & \textbf{0.430} & 1.525 & 0.829 & 0.432 \\
\textbf{Elmo+WordPiece+DWA} & \textbf{0.985} & 0.246 & 0.298 & \textbf{0.142} & \textbf{0.799} & 0.447 & 1.514 & 0.827 & \textbf{0.428}
 \\\bottomrule
\end{tabular}%
}
\caption{Experiment results on English resource test data using the monolingual models. Check \autoref{task} for word algorithm representations' abbreviation. Check \autoref{methodology} for details of monolingual models. }

\label{tab:monolingual_results}
\end{table*} 

We show monolingual models' results in Table~\ref{tab:monolingual_results}. 
As depicted, our proposed model demonstrates competitive if not the best results across the metrics. 
English, for having the most detailed dictionary data, is selected to present monolingual models' performance\footnote{ check \autoref{sec:appendix_A}}.

We notice that the transformer-based model has inferior performance on the task. 
The competition provides a low-resource data set that can explain poorer outcomes for models with high complexity. 
We tried unidirectional and bidirectional models with similar feature extractors and parameter sizes. 
The results confirm that bidirectional models perform better and benefit from grasping the context-environment more accurately.

\subsubsection{Multilingual Model Performance}

\begin{table*}[ht]
\centering
\resizebox{\textwidth}{!}{%
\begin{tabular}{@{}lccc|ccc|ccc|ccc|ccc@{}}
\toprule
\textbf{Languages}                  & \multicolumn{3}{c}{\cellcolor{L1}EN} & \multicolumn{3}{c}{\cellcolor{L2}ES} & \multicolumn{3}{c}{\cellcolor{L3}FR} & \multicolumn{3}{c}{\cellcolor{L4}IT} & \multicolumn{3}{c}{\cellcolor{L5}RU}\\ \midrule Multilingual Models & 
MSE  & COS & RANK    & MSE   & COS    & RANK  & MSE   & COS    & RANK & MSE   & COS    & RANK & MSE   & COS    & RANK  \\ \midrule
Transformer     & 1.023 & 0.201 & 0.400 & 0.977 & 0.300 & \textbf{0.310} & 1.051 & 0.278 & \textbf{0.338} & 1.143 & \textbf{0.280} & \textbf{0.340} & 0.564 & 0.318 & 0.363 \\
Transformer+RC     & 1.029 & 0.199 & 0.417 & 1.005 & 0.298 & 0.329 & 1.069 & 0.253 & 0.374 & 1.189 & 0.267 & 0.364 & 0.601 & 0.279 & 0.409 \\
Transformer+ALT     & 1.043 & \textbf{0.215} & \textbf{0.397} & 1.014 & \textbf{0.308} & 0.310 & 1.103 & \textbf{0.280} & 0.350 & 1.158 & 0.276 & 0.341 & 0.603 & \textbf{0.326} & \textbf{0.337} \\
Transformer+RC+ALT & \textbf{1.011} & 0.159 & 0.500 & \textbf{0.955} & 0.266 & 0.422 & \textbf{1.044} & 0.271 & 0.360 & \textbf{1.129} & 0.264 & 0.376 & \textbf{0.561} & 0.308 & 0.371
 \\\bottomrule
\end{tabular}%
}
\caption{Experiment results on \textbf{SGNS} word representation using the multilingual Transformer-based models. Check \autoref{methodology} for details of multilingual models. \textbf{RC} represents the Residual Cutting trick. \textbf{ALT} represents the Adding Language Token trick.}
\label{tab:multilingual_results_Transformer}
\end{table*} 

\begin{table*}[ht]
\centering
\resizebox{\textwidth}{!}{%
\begin{tabular}{@{}lccc|ccc|ccc@{}}
\toprule
\textbf{Word Representations}                  & \multicolumn{3}{c}{\cellcolor{RBRed}SGNS} & \multicolumn{3}{c}{\cellcolor{RBGreen}Char} & \multicolumn{3}{c}{\cellcolor{RBBlue}Electra} \\ \midrule Multilingual Models & 
MSE  & COS & RANK    & MSE   & COS    & RANK  & MSE   & COS    & RANK \\ \midrule  Elmo\_EN & 1.023 & 0.238 & 0.317 & 0.177 & 0.759 & \textbf{0.447} & 1.555 & 0.818 & \textbf{0.440} \\
Elmo+ALT\_EN & \textbf{1.014} & \textbf{0.246} & \textbf{0.300} & \textbf{0.164} & \textbf{0.762} & 0.449 & \textbf{1.540} & \textbf{0.825} & 0.441 \\ \midrule
Elmo\_ES & \textbf{0.953} & 0.342 & \textbf{0.234} & 0.532 & 0.810 & 0.405 & NA & NA & NA \\
Elmo+ALT\_ES & 0.960 & \textbf{0.351} & 0.235 & \textbf{0.511} & \textbf{0.822} & \textbf{0.393} & NA & NA & NA \\ \midrule
Elmo\_IT & \textbf{1.094} & 0.343 & \textbf{0.218} & 0.355 & 0.720 & 0.403 & NA & NA & NA \\
Elmo+ALT\_IT & 1.106 & \textbf{0.343} & 0.214 & \textbf{0.354} & \textbf{0.735} & \textbf{0.387} & NA & NA & NA \\ \midrule
Elmo\_FR & \textbf{1.001} & 0.313 & 0.255 & 0.388 & 0.752 & \textbf{0.411} & 1.298 & 0.845 & 0.445 \\
Elmo+ALT\_FR & 1.004 & \textbf{0.321} & \textbf{0.246} & \textbf{0.387} & \textbf{0.757} & 0.411 & \textbf{1.228} & \textbf{0.859} & \textbf{0.439} \\ \midrule Elmo\_RU & \textbf{0.547} & 0.357 & 0.247 & 0.145 & 0.816 & \textbf{0.398} & 0.891 & \textbf{0.729} & 0.386 \\
ELmo+ALT\_RU & 0.563 & \textbf{0.368} & \textbf{0.232} & \textbf{0.137} & \textbf{0.828} & 0.400 & \textbf{0.887} & 0.728 & \textbf{0.384}
 \\\bottomrule
\end{tabular}%
}
\caption{Experiment results of the multilingual ELmo-based models. \textbf{ALT} represents the Adding Language Token trick.}
\label{tab:multilingual_results_Elmo}
\end{table*}

We show two ablation experiment results to explain the influence of adding language tags and residual connection removal. 
\textbf{First}, experiment results of the Transformer-based multilingual model on SGNS embedding can suggest the benefits of language tags and curbing residual connection separately or jointly.
\textbf{Second}, we propose experimental results of the original and adjusted Elmo-based multilingual models. The latter subsumes added language tokens. Such a comparison would clarify whether adding language tokens lead to a general improvement across different languages and word representations. 

Electra word representations of Spanish and Italian are not available, implying no related experimental results.
The outcomes demonstrate that multilingual models benefit from language-specific information but not from language-agnostic structure. 
Adding language tags has proved a positive influence on various language models.

\subsection{Ablation Study}

\subsubsection{Tokenizer}

We tried three widely-used tokenizers for our proposed model: BPE, ULM, and WordPiece. 
Both ULM and WordPiece show competitive performance in transformer- and Elmo-based structures. 
BPE has relatively low performance since the data resource is insufficient and has higher resource requests.

\subsubsection{Multitask Model}

According to the performance comparison in Table~\ref{tab:monolingual_results}, DWA helps the Elmo model achieve better performance and reconstructs three-word representations simultaneously. 
It demonstrates that differently learned word representations have an internal relation and can be learned together using a shared bottom structure.

\subsubsection{Difficulty of Reconstructing Different Word Representations}

Compared with the Char and Electra, we find that the SGNS is harder to learn from the gloss corpus, suggesting that the contextualized information of words in sentences might be missing from the pure dictionary glosses. 
Additionally, the result along with \cite{kaneko2021dictionary} indicates dictionary corpus can be a promising way to remove the unfair biases rooted in large corpus learned word embeddings.

\subsubsection{Difficulty of Learning Different Languages}

\begin{table}[ht]
\centering
\caption{Language Vocabulary Size Ablation Study. \textbf{Dict. Size} means the number of non-repeating tokens shown in the glosses. \textbf{Avg. Gloss Len} means the average token numbers contained in a gloss.}
\resizebox{0.45\textwidth}{!}{%
\begin{tabular}{@{}lc|c|c|c@{}}
\toprule
\textbf{Languages}                  
& \multicolumn{1}{c}{Gloss Num} 
& \multicolumn{1}{c}{Dict.Size} 
& \multicolumn{1}{c}{Avg.Gloss Len} 
& \multicolumn{1}{c}{Elmo SGNS COS}
\\
\midrule
English &43608	&29042	&11.7 &0.252 \\ 
French &43608	&40028	&14.3 &0.333 \\ 
Italian &43608	&40126	&13.6 &0.352 \\
Spanish &43608	&46761	&14.8 &0.362 \\ 
Russia &43608   &57137	&11.3 &0.387 \\
\bottomrule
\end{tabular}%
}
\label{tab:table_dictionary}
\end{table}

Our results of experiments show a strong positive correlation between language's tokens dictionary size and the models' achievable performance \autoref{tab:table_dictionary}. 

There are several possible reasons for the observation. 
First, as the language model dictionary size decreases, the models' and glosses' ability to explain the slight differences between words, especially the polysemies and synonyms, decreases. 
Second, a smaller dictionary size indicates that the covered tokens in the language model are a relatively incomplete part of words of the language.

\textbf{Noted} that the second explanation above does not consider the intrinsic differences between languages. The morphologically rich languages, like Russian, tend to have larger vocabulary sizes and bring many unknown words that influence performance negatively \cite{jurafsky_2020_speech}.

\section{Conclusion}

The paper proposes a model showing competitive results in most cases of the reverse dictionaries task. Several conclusions are provided about the reverse dictionaries task by the paper based on the ablation studies. \textbf{First}, the transformer-based model, for its high complexity, performs worse compared to RNN- or LSTM-based models. Multilingual transformer-based model benefits from specifying languages and including language-related grammar positional information. \textbf{Second}, bidirectional models with similar parameter sizes outperform the unidirectional one since they better grasp the context in low-resource conditions. \textbf{Third}, different word representations are potential connections and can be collaboratively learned from glosses using a multitask learning structure. SGNS embedding is much harder to model compared to Character embedding and Electra embedding.

\section{Acknowledgements}

We express our gratitude to Prof. Shi Wang and Prof. Alexis Palmer for providing computing resources and guidance. 
We are grateful to the organizers for providing such a fascinating and inspiring competition and promptly resolving all our questions. 
Special thanks to Rebecca Lee, Xingran Chen, and Natalia Wojarnik for idea sharing and a deep discussion in the initial stage.

\appendix

\section{Appendix:A}

Check Table~\ref{tab:appendix_A} for the Appendix A.
 
\label{sec:appendix_A}

\begin{table*}[ht]
\centering
\caption{\textbf{Appendix A}. Experiment results of the monolingual models. Check \autoref{task} for word algorithm representations' abbreviation. Check \autoref{methodology} for details of monolingual models. }
\resizebox{\textwidth}{!}{%
\begin{tabular}{@{}lccc|ccc|ccc@{}}
\toprule
\textbf{Word Representations}                  & \multicolumn{3}{c}{\cellcolor{RBRed}SGNS} & \multicolumn{3}{c}{\cellcolor{RBGreen}Char} & \multicolumn{3}{c}{\cellcolor{RBBlue}Electra} \\
\midrule Monolingual Models & 
MSE  & COS & RANK    & MSE   & COS    & RANK  & MSE   & COS    & RANK  \\ \midrule &  \multicolumn{9}{c}{\cellcolor{RBBlue2}\textbf{Language} English} \\ \midrule
RNN+WordPiece     & 1.000   & 0.249   & 0.310   & 0.158 & 0.778 & 0.442 & \textbf{1.454} & \textbf{0.832} &\textbf{0.433}   \\
LSTM+WordPiece &\textbf{0.990}	 & 0.228 & 0.375&\textbf{0.148} &\textbf{0.791} & 0.458 & 1.491 & 0.831 & 0.449 \\
Transformer+WordPiece & 1.042 & 0.214 & 0.367 & 0.194 & 0.780 & 0.453 & 1.796 & 0.827 & 0.486 \\
BiRNN+WordPiece & 0.989 & 0.221 & 0.395 & 0.150 & 0.791 & 0.454 & 1.483 & 0.832 & 0.449 \\
Elmo+WordPiece & 1.041 &\textbf{0.252} &\textbf{0.282} & 0.161 & 0.772 &\textbf{0.430} & 1.512 & 0.829 & 0.434 \\ 

\midrule  &  \multicolumn{9}{c}{\cellcolor{RBBlue2}\textbf{Language} Spanish} \\ \midrule
RNN+WordPiece  &0.936 &0.358 &0.225 &0.512 &0.822 &0.402 &NA  &NA &NA   \\
LSTM+WordPiece &\textbf{0.928}	&0.334	&0.287	&\textbf{0.497}	&\textbf{0.829}	&0.418 &NA  &NA &NA \\
Transformer+WordPiece &1.011	&0.307	&0.313	&0.577	&0.828	&0.432 &NA  &NA &NA \\
BiRNN+WordPiece &0.939	&0.315	&0.329	&0.511	&0.826	&0.423 &NA  &NA &NA \\ 
Elmo+WordPiece &0.968	&\textbf{0.362}	&\textbf{0.207}	&0.520	&0.820	&\textbf{0.396} &NA  &NA &NA   \\

\midrule  &  \multicolumn{9}{c}{\cellcolor{RBBlue2}\textbf{Language} French} \\ \midrule 
RNN+WordPiece  &0.975	&0.329	&0.254	&0.379	&0.761	&0.408	&1.272	&0.856	&0.444   \\
LSTM+WordPiece &\textbf{0.971}	&0.303	&0.329	&\textbf{0.361}	&\textbf{0.772}	&0.420	&\textbf{0.191}	&0.862	&0.457 \\
Transformer+WordPiece &1.057	&0.273	&0.366	&0.461	&0.771	&0.430	&1.523	&0.856	&0.488   \\
BiRNN+WordPiece &0.984	&0.290	&0.361	&0.366	&0.770	&0.424	&1.202	&\textbf{0.863}	&0.454 \\
Elmo+WordPiece &1.007	&\textbf{0.333}	&\textbf{0.239}	&0.373	&0.763	&\textbf{0.402}	&1.341	&0.850	&\textbf{0.437} \\ 

\midrule  &  \multicolumn{9}{c}{\cellcolor{RBBlue2}\textbf{Language} Italian} \\ \midrule
RNN+WordPiece  &1.078	&\textbf{0.353}	&0.218	&0.345	&0.741	&0.391  &NA  &NA &NA\\
LSTM+WordPiece &\textbf{1.077}	&0.324	&0.276	&0.340	&0.744	&0.413 &NA  &NA &NA\\
Transformer+WordPiece &1.160	&0.256	&0.373	&0.377	&0.731	&0.419  &NA  &NA &NA \\
BiRNN+WordPiece &1.086	&0.309	&0.303 &\textbf{0.338}	 &\textbf{0.747} &0.415 &NA  &NA &NA\\
Elmo+WordPiece &1.106	&0.352	 &\textbf{0.200}	&0.354	&0.736	&\textbf{0.384} &NA  &NA &NA\\

\midrule  &  \multicolumn{9}{c}{\cellcolor{RBBlue2}\textbf{Language} Russian} \\ \midrule
RNN+WordPiece  &\textbf{0.537} &\textbf{0.388}	&0.226	&0.132	&0.832	&0.391	&0.899	&0.727	&0.372   \\
LSTM+WordPiece &0.547	&0.338	&0.346	&\textbf{0.131}	&\textbf{0.834}	&0.401	&\textbf{0.885}	&\textbf{0.728}	&0.400 \\
Transformer+WordPiece &0.565	&0.315	&0.377	&0.156	&0.827	&0.411	&1.071	&0.707	&0.473   \\
BiRNN+WordPiece &0.551	&0.321	&0.397	&0.135	&0.831	&0.403	&0.919	&0.727	&0.410 \\
Elmo+WordPiece &0.557	&0.387	&\textbf{0.217}	&0.134	&0.831	&\textbf{0.390}	&0.904	&0.723	&\textbf{0.362} \\ 

\bottomrule
\end{tabular}%
}
\label{tab:appendix_A}
\end{table*}

\section{Appendix:B}
\label{sec:appendix_B}

Check Table~\ref{tab:appendix_B} for the Appendix B.

\begin{table*}[ht]
\centering
\caption{\textbf{Appendix B}. The table shows the selected multilingual models' performance. Check \autoref{task} for word algorithm representations' abbreviation. Check \autoref{methodology} for details of monolingual models. }
\resizebox{\textwidth}{!}{%
\begin{tabular}{@{}lccc|ccc|ccc@{}}
\toprule
\textbf{Word Representations}                  & \multicolumn{3}{c}{\cellcolor{RBRed}SGNS} & \multicolumn{3}{c}{\cellcolor{RBGreen}Char} & \multicolumn{3}{c}{\cellcolor{RBBlue}Electra} \\
\midrule Monolingual Models & 
MSE  & COS & RANK    & MSE   & COS    & RANK  & MSE   & COS    & RANK  \\ \midrule &  \multicolumn{9}{c}{\cellcolor{RBBlue2}\textbf{Language} English} \\ \midrule
Elmo+WordPiece & 1.041 &\textbf{0.252} &\textbf{0.282} & 0.161 & 0.772 &\textbf{0.430} &\textbf{1.512} &\textbf{0.829} & 0.434 \\ 
Elmo + WordPiece + DWA &\textbf{0.985}	&0.246	&0.298	&\textbf{0.142}	&\textbf{0.799}	&0.447	&1.514	&0.827	&\textbf{0.428}\\

\midrule &  \multicolumn{9}{c}{\cellcolor{RBBlue2}\textbf{Language} French} \\ \midrule
Elmo+WordPiece  &1.007	&\textbf{0.333}	&\textbf{0.239}	&0.373	&0.763	&\textbf{0.402}	&1.341	&0.850	&0.437 \\ 
Elmo + WordPiece + DWA &\textbf{0.937}	&0.327	&0.243	&\textbf{0.364}	&\textbf{0.770}	&0.406	&\textbf{1.315}	&\textbf{0.854}	&\textbf{0.428}\\

\midrule &  \multicolumn{9}{c}{\cellcolor{RBBlue2}\textbf{Language} Russian} \\ \midrule
Elmo+WordPiece  &0.557	&0.387	&0.217	&0.134	&0.831	&\textbf{0.390}	&0.904	&0.7226	&\textbf{0.362} \\ 
Elmo + WordPiece + DWA &\textbf{0.534}	 &\textbf{0.388}	 &\textbf{0.189}	 &\textbf{0.127}	 &\textbf{0.838}	&0.376	 &\textbf{0.908}	 &\textbf{0.7235}	&0.364\\
\bottomrule
\end{tabular}%
}
\label{tab:appendix_B}
\end{table*}

\bibliographystyle{acl_natbib}
\bibliography{custom}

\end{document}